%% file: main.tex
\documentclass[sigconf, nonacm, pdfa]{acmart}

\newcommand\vldbdoi{10.14778/3773731.3773740}
\newcommand\vldbpages{5652 - 5661}
\newcommand\vldbvolume{18}
\newcommand\vldbissue{13}
\newcommand\vldbyear{2025}
\newcommand\vldbauthors{\authors}
\newcommand\vldbtitle{\shorttitle} 
\newcommand\vldbavailabilityurl{https://github.com/swjz/data-minimization}
\newcommand\vldbpagestyle{empty}
\usepackage{tikz}

\usepackage{filecontents}

\usepackage{graphicx}
\usepackage{caption}
\usepackage{subcaption}
\usepackage{booktabs} 

\usepackage{amssymb}
\usepackage{listings}
\usepackage{array}
\usepackage{xspace} 
\usepackage[pdf]{graphviz}
\usepackage[normalem]{ulem} 
\usepackage{comment}
\graphicspath{ {./figures/} }
\usepackage{varwidth}
\usepackage[ruled, linesnumbered]{algorithm2e}
\usepackage{enumitem}
\usepackage{footnote}
\usepackage{multirow}
\usepackage{tablefootnote}
\usepackage{xpatch}
\usepackage{algorithm2e}
\usepackage{balance}
\usepackage{makecell}
\usepackage{amsmath}
\usepackage{amsthm}
\theoremstyle{definition}  
\newtheorem{definition}{Definition}
\newtheorem{assumption}{Assumption}
\usepackage[normalem]{ulem}
\usepackage{bm}
\usepackage{placeins}
\usepackage[final, commandnameprefix=ifneeded]{changes}

\definechangesauthor[color=red]{Scope}
\definechangesauthor[color=green]{Typo}
\definechangesauthor[color=blue]{Design}
\definechangesauthor[color=magenta]{Organization}
\definechangesauthor[color=orange]{Evaluation}
\definechangesauthor[color=brown]{Transfer}
\definechangesauthor[color=purple]{Clarification}
\definechangesauthor[color=olive]{Discussion}

\usepackage{listings}
\usepackage[a-2b,mathxmp]{pdfx}

\newcolumntype{$}{>{\global\let\currentrowstyle\relax}}
\newcolumntype{^}{>{\currentrowstyle}}
\newcommand{\rowstyle}[1]{\gdef\currentrowstyle{#1}%
  #1\ignorespaces
}
\newcommand{\emoji}[1]{\includegraphics[width=1em]{figures/emoji/#1}}

\begin{document}
\title{Algorithmic Data Minimization for Machine Learning over Internet-of-Things Data Streams}

\author{Ted Shaowang}
\affiliation{%
  \institution{The University of Chicago}
}
\email{swjz@uchicago.edu}

\author{Shinan Liu}
\affiliation{%
  \institution{The University of Hong Kong}
}
\email{shinan6@hku.hk}

\author{Jonatas Marques}
\affiliation{%
  \institution{The University of Chicago}
}
\email{jmarques@uchicago.edu}

\author{Nick Feamster}
\affiliation{%
  \institution{The University of Chicago}
}
\email{feamster@uchicago.edu}

\author{Sanjay Krishnan}
\affiliation{%
  \institution{The University of Chicago}
}
\email{skr@uchicago.edu}


\begin{abstract}
Machine learning can analyze vast amounts of data generated by IoT devices to identify patterns, make predictions, and enable real-time decision-making.
This raises significant privacy concerns, necessitating the application of data minimization -- a foundational principle in emerging data regulations, which mandates that service providers only collect data that is directly relevant and necessary for a specified purpose.
Despite its importance, data minimization lacks a precise technical definition in the context of sensor data, where collections of \textit{weak signals} make it challenging to apply a binary ``relevant and necessary'' rule.
This paper provides a technical interpretation of data minimization in the context of sensor streams, explores practical methods for implementation, and addresses the challenges involved.
Through our approach, we demonstrate that our framework can reduce user identifiability by up to 16.7\% while maintaining accuracy loss below 1\%, offering a viable path toward privacy-preserving IoT data processing.
\end{abstract}

\maketitle



\pagestyle{\vldbpagestyle}
\begingroup\small\noindent\raggedright\textbf{PVLDB Reference Format:}\\
\vldbauthors. \vldbtitle. PVLDB, \vldbvolume(\vldbissue): \vldbpages, \vldbyear.\\
\href{https://doi.org/\vldbdoi}{doi:\vldbdoi}
\endgroup
\begingroup
\renewcommand\thefootnote{}\footnote{\noindent
This work is licensed under the Creative Commons BY-NC-ND 4.0 International License. Visit \url{https://creativecommons.org/licenses/by-nc-nd/4.0/} to view a copy of this license. For any use beyond those covered by this license, obtain permission by emailing \href{mailto:info@vldb.org}{info@vldb.org}. Copyright is held by the owner/author(s). Publication rights licensed to the VLDB Endowment. \\
\raggedright Proceedings of the VLDB Endowment, Vol. \vldbvolume, No. \vldbissue\ %
ISSN 2150-8097. \\
\href{https://doi.org/\vldbdoi}{doi:\vldbdoi} \\
}\addtocounter{footnote}{-1}\endgroup

\ifdefempty{\vldbavailabilityurl}{}{
\vspace{.3cm}
\begingroup\small\noindent\raggedright\textbf{PVLDB Artifact Availability:}\\
The source code, data, and/or other artifacts have been made available at \url{\vldbavailabilityurl}.
\endgroup
}

\input{introduction}

\input{background}
\input{system}

\input{evaluation}
\input{discussion}
\balance

\bibliographystyle{ACM-Reference-Format}
\bibliography{reference}

\end{document}

%% file: introduction.tex
\section{Introduction}
Internet of Things (IoT) systems consist of interconnected devices that collect and transmit data through embedded sensors, enabling smart functionalities across various domains. Machine learning plays a critical role in analyzing this sensor data to unlock valuable insights and optimize operations~\cite{iotml1, iotenergy, iothealth,liu2023leaf,shaowang2023amir}. For example, in smart homes, machine learning models can predict energy consumption patterns to optimize heating or cooling schedules, reducing costs and environmental impact. In healthcare, IoT devices, such as wearable sensors, can track vital signs and alert medical professionals to anomalies, enabling proactive care. These applications highlight the vast opportunities IoT and machine learning offer~\cite{wan2024cato,jiang2023ac}, from enhancing personalized experiences to improving efficiency and fostering innovation across industries.

However, with these opportunities come certain acute risks. IoT data, e.g., from smart home systems, can inadvertently leak identifiable information due to vulnerabilities in data handling and transmission~\cite{iotsec1, iotsec2, iotsec3}. Devices such as smart speakers, thermostats, and security cameras collect and transmit vast amounts of personal data, including voice commands, daily routines, and even video footage. Such risks highlight the critical need for robust measures and transparency in IoT ecosystems to protect user data and privacy. Thus, every service provider building an intelligent IoT system must tradeoff the utility of collecting data for prediction versus the potentially identifiable characteristics manifested in the data~\cite{iotsec4,yang2025towards}. 

To help navigate such tradeoffs, emerging frameworks like the General Data Protection Regulation (GDPR) and the California Consumer Privacy Act (CCPA) have introduced new principles for managing user data~\cite{GDPR2016a,ccpa2018}. Among these, \textit{data minimization} stands out as a foundational principle.
It mandates that service providers collect only the user data that is directly ``relevant and necessary'' to achieve a specified purpose~\cite{datamini1, datamini2, datamini3, iotsec4}.
The principle of data minimization is important because it reduces the risk of data breaches, ensures compliance with privacy regulations, and limits the collection of unnecessary or excessive personal information, thereby protecting individual privacy and fostering trust~\cite{datamini1}.

Unfortunately, a clear technical definition of data minimization is missing in the context of IoT analytics. In a sensor setting, a threshold of ``relevant and necessary'' may be hard to characterize as every data facet may be correlated in some way to the desired prediction target. In these settings, we often have collections of weak signals that contribute both towards a prediction target and towards re-identifying individuals.
Determining what data is truly necessary remains an ill-posed task.

In this work, we address this challenge by proposing a novel approach to data minimization. Our method focuses on selectively discarding features that, while potentially useful for identifying individual users, offer minimal contribution to the task at hand. This strategy aims to strike a balance between privacy preservation and task accuracy, ensuring that privacy-sensitive information is minimized without significantly compromising system performance.
It is crucial to clarify that data minimization does not equate to simply minimizing the size of the collected dataset (i.e., rows). \added[id=Scope]{We focus on feature minimization to address data leakage during inference time.} Rather, the goal is to minimize \textit{identifiability} of users inferred from the dataset based on the features used for a machine learning task. 

This paper makes the following contributions.
(1) A formal model for data minimization based on a two-player game in which a model provider tries to maximize accuracy while an adversary tries to maximize identifiability of users.
(2) We present practical algorithms to identify effective provider strategies as solutions to this game, whose optimal solution is computationally intractable.
(3) We show experimental results across 7 IoT datasets. While no single heuristic is universally effective, we show that data minimization strategies that \emph{do not} model this two-player game with an adversary are generally less effective. 
(4) We provide concrete recommendations on best practices to improve the effectiveness of data minimization in real-world applications.

%% file: background.tex
\section{Background}
Machine learning over multimodal IoT data involves integrating information from diverse sensor sources—such as cameras, accelerometers, and biosensors—to improve decision-making, prediction accuracy, and situational awareness. These sensors often capture complementary aspects of an environment, enabling more robust models for applications like autonomous vehicles, healthcare monitoring, and industrial automation. For example, in medical diagnostics, combining EEG (brain activity) and ECG (heart activity) data can enhance early detection of neurological and cardiovascular disorders. Multimodal sensing is characterized by fusing information from multiple ``weak'' signals -- rather than a single strong one like in classical AI problems such as NLP and Computer Vision. Sensor data is plagued with missing or degraded signals, which can be mitigated through multiple sensing views of the same phenomenon.

Unfortunately, the nature of IoT applications means that the data collected could leak identifying information.
Consider the EEG application above.
While clinically useful to share EEG datasets, these data can inadvertently reveal sensitive information. 
Idiosyncrasies in sensing hardware or the individual could leak data to an adversary.
This issue is particularly pronounced in cases where specific signals are unique to certain users. 
To address this, privacy-preserving frameworks are needed to help data and model providers navigate accuracy vs. identifiability tradeoffs.

\subsection{Formal Approaches to Data Minimization}
The principle of data minimization is a fundamental concept in data privacy and security that emphasizes collecting, processing, and storing only the minimum amount of personal data necessary to achieve a specific purpose. This principle is a core requirement in many data protection regulations, such as the General Data Protection Regulation (GDPR) and the California Consumer Privacy Act (CCPA). By limiting data collection, organizations reduce the risk of data breaches, unauthorized access, and misuse while also ensuring compliance with legal and ethical standards.
 
Unfortunately, the IoT domain introduces specific challenges. It isn't a clear yes/no question to whether a particular signal is absolutely needed for a task or is overly identifying. Every signal has a degree of utility as well as a degree of identifiability. This degree of utility can be highly dynamic, varying across the feature space and over users. This paper explores formal approaches to data minimization in such applications.

Data minimization can take various forms, including the removal of data points (rows) and features (columns). 
\added[id=Scope]{Removing rows as a form of data minimization is highly valuable for ensuring training datasets do not contain too much private information. Conversely, feature-minimization largely helps with reducing the amount of data given to an inference service~\cite{NEURIPS2023_e48880ea}. For the purpose of this paper, we focus exclusively on removing unnecessary features (columns) for optimizing the inference problem. Both approaches are complementary as they address different stages of the machine-learning pipeline.}
This approach is comparable to traditional feature selection methods, as discussed in \S\ref{sec:feature-selection}. However, traditional methods typically prioritize utility by selecting features that maximize model accuracy. They overlook the privacy implications of certain features, which may inadvertently reveal user identity.
We argue that an effective data minimization method should strike a balance between utility and privacy~\cite{privacy-utility}. It should retain features that are essential for the predictive task while removing those that reveal user information, provided their contribution to accuracy is minimal.

A key gap in existing work is the lack of a clear definition of user identifiability -- what does it mean for a set of features to reveal user information? We propose a practical definition: user identifiability can be measured by training a user-classification model on such features. A privacy-preserving set of features should result in low accuracy for this model, indicating that user information cannot be inferred from these features.


\begin{definition}[Data Minimization]~\label{def:data-min-formal}
The goal of data (feature) minimization is to find a subset $S$ of feature set $F$ $(S\subseteq F)$:
\[
\text{min Identifiability}[t(S)]
\]
subject to:
\[
\text{Accuracy}[p(S)] \geq (1-l) \times \text{Accuracy}[p(F)]
\]
where $p$ is the primary model for the predictive task, and $t$ is a user-classification model. Identifiability is defined as the classification accuracy of $t$.
\end{definition}
When certain features strongly influence both predictive accuracy and user identifiability, the tradeoff between utility and privacy becomes more nuanced. \added[id=Scope]{The extent to which users are willing to sacrifice predictive performance for enhanced privacy is controlled by the tunable tolerance parameter $l$.
This threshold plays a critical role in practice, as users may be unwilling to compromise accuracy arbitrarily in favor of privacy.}

\subsection{Related Work and Baselines}\label{sec:baselines}
Existing work mostly approaches privacy-aware data management problems from one of the three perspectives: (1) selection strategy, (2) privacy mechanism, and (3) computation location.
Privacy mechanisms include differential privacy~\cite{ALISHAHI2022102934}, homomorphic encryption~\cite{a15070229} and secure multi-party computation~\cite{10.1007/978-3-031-70890-9_18}.
Computation locations include centralized, federated learning~\cite{pmlr-v202-pang23a} and edge-based serving~\cite{shaowang2023edgeserve,shaowang2021declarative,shaowang2022bidede}.
In this section, we mainly focus on selection strategies as they are more relevant to our work.



\subsubsection{Differential privacy}
Differential privacy is a well-studied technique to anonymize private data by injecting calibrated noise to a dataset while preserving its overall statistics~\cite{10.1007/11681878_14}. This method ensures that individual data cannot be identified by a statistical analysis or release.
While differential privacy makes it harder to reveal information about a specific user, it does not take into consideration what data is relevant and necessary for the intended task.
This limitation can lead to the retention of data that is irrelevant to the task and accuracy loss.

\subsubsection{Feature selection}\label{sec:feature-selection}
Existing feature selection methods focus on selecting a subset of features that are most significant to the model's performance.
Although this process does reduce the total amount of information, it overlooks the possibility that the most predictive features might also be the most privacy-intrusive.
Consequently, these methods may inadvertently retain features that pose a higher risk to user identifiability, despite their contribution to the model's accuracy.

\noindent \textbf{Feature hashing.}
The hashing trick~\cite{10.5555/2969735.2969739, 10.1145/1553374.1553516} can be used as a tool for feature selection as it hashes a large number of features into a smaller number of indices.
A fixed-size output dimensionality is guaranteed, but hash collisions can occur and add noise to data.

\noindent \textbf{PCA-based methods.}
PCA~\cite{Pearson01111901} and Sparse PCA~\cite{zou2006sparse} are examples of linear methods to reduce the dimensionality of features.
For dense datasets, PCA would work just fine but Sparse PCA works better when the dataset is high-dimensional and sparse.

\subsubsection{Feature scoring.}\label{sec:feature-scoring}
Another line of work discusses how to attribute scores to each feature so it is easier to determine the contributions made by each feature.

\noindent \textbf{Entropy-based methods.}
Mutual information is a measure of shared information between two variables, \added[id=Clarification]{or the entropy lost by knowing one of the variables.}~\cite{kreer1957question}.
A high mutual information score means that knowing $X$ reduces more uncertainty about $Y$.
In the context of feature selection, we want to keep the features that share high mutual information with labels.

Following ideas from~\cite{privacy-utility}, we apply mutual information as a metric for feature \textit{utility score}, and \added[id=Clarification]{the entropy difference between including the feature vs. excluding the feature as a metric for feature \textit{privacy score}. This is just the conditional entropy of the included feature given all others, i.e., how identifying to an arbitrary labeling is this feature above what is already in the dataset.}
The normalized sum of utility score and privacy score can be used as a \textit{privacy-utility tradeoff score}.
However, we find that high entropy features do not necessarily mean privacy-invasive in many cases, as long as the additional entropy does not correlate with personal identity.


\noindent \textbf{SHAP-based methods.}\label{sec:shap}
SHAP (SHapley Additive exPlanations)~\cite{NIPS2017_7062} is a game-theoretic approach to explain machine learning models by attributing changes in model outputs to each contributing feature based on Shapley values.

\noindent \textbf{Impurity-based feature importance.}
Gini impurity is yet another metric for feature selection, particularly in tree-based models such as random forests and gradient boosted trees~\cite{gini1912variabilita}.
This method evaluates the importance of a feature based on its contribution to reducing impurity in the decision nodes of the model.
Features that lead to greater reductions in impurity across multiple splits are considered more informative and receive higher importance scores.


%% file: system.tex
\section{Formal Model for Data Minimization Over Weak Signals}\label{sec:formal-model}
Next, we present a method for operationalizing data minimization over practical IoT machine-learning applications.

\noindent \textbf{Notation: }
First, we define the following notation and terms.
\begin{itemize}
    \item \textbf{Features}. Each feature $f$ in a feature set $F$ represents a distinct signal (modality) that can be used by the model. Features can be single-dimensional (scalar) or more complex.

    \item \textbf{Labeling}. A labeling is an assignment of each training example $x_i \in X$ in a dataset to a label $y_i \in Y$. Labels can be categorical, real-valued, or vector-valued.

    \item \textbf{Model}. A model is a predictor that can take as input $F' \subseteq F$ to predict some target label. That is, it is trained on a projection of $X[F']$ to predict some $Y$.
\end{itemize}

\subsection{Re-identification Game}\label{sec:game}
We formalize the data minimization problem as a two-player game consisting of a \textsf{Provider} and an \textsf{Adversary}. As the names imply, \textsf{Provider} is trying to solve some IoT task with machine learning and the \textsf{Adversary} is trying to re-identify users. 

This game can be modeled as a hypothetical optimization that happens at training time.
Both players have knowledge of the training dataset $X$ which has two labelings $Y_{task}$ and $Y_{user}$. $Y_{task}$ describes a desired prediction target of the \textsf{Provider}. $Y_{user}$ describes a user/entity/individual identifier the \textsf{Provider} is trying to hide. There is a preset parameter $\ell$ that both players know and an oracle that measures the accuracy of any model $acc(m)$. The game proceeds as follows:

\begin{enumerate}
    \item The \textsf{Provider} selects a subset of features $F'$ and presents a model $m$ that predicts $Y_{task}$ from the subset.
    \begin{itemize}
        \item If $acc(m) < \ell$, the \textsf{Provider} automatically loses the game; a reward score of $-\infty$.
    \end{itemize}
    \item The \textsf{Adversary} observes $F'$ and $m$, and presents a model $m_{adv}$ that predicts $Y_{user}$ from the same subset of features.
    \item The \textsf{Provider} receives a reward score $- acc(m_{adv})$; the better the user-classification model, the worse the score.
\end{enumerate}

In this game, the \textsf{Provider} is trying to maximize their reward score, which means finding a subset of features that does not degrade accuracy beyond $\ell$ (which can be calibrated against a model trained on the entire set $F$) while minimizing the accuracy of a potential adversary (i.e. identifiability). To be able to tractably identify a strategy for the \textsf{Provider}, we need to make a simplifying assumption:

\begin{assumption}[Model Class Parity]\label{model-class-parity}
The \textsf{Adversary} chooses a model $m'$ from the same pre-defined model family as the $m$ given by the provider.
\end{assumption}

In other words, if the \textsf{Provider} can define a broad class of acceptable models as a part of their training procedure, e.g., random forests or linear models. $m$ is the best model in that class that predicts $Y_{task}$ that is yielded from parameter and hyperparameter optimization. Assumption~\ref{model-class-parity} enforces that $m'$ is found by the same procedure but against a different labeling. This assumption enforces a few desirable properties: (1) the \textsf{Adversary} is not substantially more capable than the \textsf{Provider} at prediction, (2) there is a bounded world of techniques that the \textsf{Adversary} can use, and (3) the \textsf{Adversary} is implicitly assuming that the \textsf{Provider} is a subject matter expert that has picked the best model for this feature space. 

Of course, the above game is NP-Hard to solve, even in the simplified setting where 
each feature $f$ has an identifiability score (cost) $c_f$ and a utility score (value) $v_f$. Assuming that the utility and identifiability of each feature $f$ are linearly additive, the optimal solution reduces to a 0-1 Knapsack problem:
\[
\text{min}\sum _{f\in F}c_f \qquad \text{s.t.}\qquad \sum _{f\in F}v_f\geq V
\]

where $V$ is a total utility threshold that controls the number of features selected. 
In practice, decomposing a feature space into identifiability and utility scores is very challenging. Complex ML problems often have multi-feature interactions, where the value of one feature is conditioned on the presence or absence of another.

\subsection{Provider Strategy as an API}\label{sec:api}
The rest of this paper shows how we turn the re-identification game into a feature optimization API that providers can use to achieve some data minimization.
We assume that the service provider makes a best effort to protect user privacy; however, certain features may still act as side channels, leaking information that can be used to infer user identity.
Thus, the optimal provider strategy from the game above gives a provider guidance on how to navigate this trade-off.
The API is designed as follows:

\begin{lstlisting}
def feature_minimize(features, utility_labels, user_labels, threshold) -> minimized_features
\end{lstlisting}

The user-defined threshold $l$ (which is a calibrated version of $\ell$ above) is a number between 0 and 1, representing how much accuracy loss the user is willing to tolerate while optimizing for lower identifiability. Setting the threshold to 1 means that the user absolutely wants the lowest identifiability, even if the accuracy reaches 0\%; a threshold of 0 means that no accuracy loss is acceptable, and the algorithm can only search for feature subsets that maintain the same accuracy as the original dataset. For instance, a threshold of 0.1 implies that the user is comfortable with 10\% accuracy loss compared to its original level. The algorithm will then search for feature subsets as long as the accuracy remains at least 90\% of the level before data minimization.

\begin{definition}[Relative Effectiveness]
\added[id=Scope]{
We define the relative effectiveness $r_i$ as follows:
\[
r_i = \log (\max(\frac{\text{Identifiability}_0 - \text{Identifiability}_i}{\text{Accuracy}_0 - \text{Accuracy}_i}, \epsilon))
\]}
where $\text{Identifiability}_i$ and $\text{Accuracy}_i$ represent the identifiability and accuracy at threshold $i$ while $\text{Identifiability}_0$ and $\text{Accuracy}_0$ correspond to their values at threshold 0.
\added[id=Scope]{This threshold corresponds to the parameter $l$ defined in Definition~\ref{def:data-min-formal}.
We include a small constant $\epsilon>0$ to clip the ratio.}
This metric quantifies how sensitive identifiability is to changes in accuracy. A higher $r_i$ indicates that identifiability can be significantly reduced with minimal accuracy loss, making the trade-off more favorable.
\end{definition}

\section{Practical Solutions to the Re-Identification Game}\label{sec:practical-solutions}
Behind our API lies a practical algorithm implementing the \textsf{Provider} strategy. We begin with an optimal but computationally inefficient approach~(\S\ref{sec:brute-force}), followed by a more efficient but suboptimal alternative~(\S\ref{sec:greedy}), and finally a hybrid solution that balances computational feasibility and optimality by combining the strengths of both~(\S\ref{sec:hybrid}).
\subsection{Simple Exhaustive Search}\label{sec:brute-force}
A straightforward strategy for the \textsf{Provider} is to perform an exhaustive search on all possible subsets of the feature set $F$.
This involves training a primary model $m$ and a user-classification model $m_{adv}$ on each of the $2^{|F|}$ subsets.
Then, we log all the accuracies for the primary model and the identifiabilities for the user-classification model. All these feature subsets are ranked based on the following priorities:
\begin{enumerate}
  \item Identify the highest achievable primary model accuracy among all feature subsets and gradually relax it according to the user-defined threshold $l$.
  \item Filter out feature subsets that do not meet the specified accuracy threshold.
  \item Sort the remaining feature subsets in ascending order of user-classification model identifiability, prioritizing those with the lowest identifiability.
The top result is our desired feature set with the best utility-identifiability tradeoff.
\end{enumerate}
Although this approach guarantees optimality according to Definition~\ref{def:data-min-formal}, it is computationally infeasible for large $|F|$. Even though the process is perfectly parallelizable, exhaustive search on more than 15 features can take hours on a commodity server, making it impractical for high-dimensional datasets.

\subsection{Greedy Selection}\label{sec:greedy}
A more efficient but less optimal alternative to exhaustive search is to assume that each feature's contribution is linearly additive. Under this assumption, we can decompose accuracy and identifiability into per-feature values, allowing us to approximate feature importance without evaluating all possible subsets.
The key idea is to develop a scoring mechanism that quantifies each feature’s impact on both:
\begin{itemize}
    \item The primary model’s accuracy (utility score, or value).
    \item The user-classification model’s accuracy (identifiability score, or cost).
\end{itemize}
As described in~\ref{sec:feature-scoring}, several techniques can be used to estimate these scores.
For example, we can apply the mutual information-based utility score and the entropy-based privacy score from~\cite{privacy-utility} as our utility score and identifiability score here, respectively.
Similarly, we can take the Gini impurity of both models and apply the feature importance values to each feature as utility and identifiability scores.
As an alternative, we can also calculate the SHAP value for both models and represent them as scores as follows:

Let $\bm{S_{adv}}$ be the SHAP values for the user-classification model along the feature axis, and $\bm{S}$ be the SHAP values for the primary model along the feature axis.
\begin{itemize}
    \item The utility score $\bm{v}=\text{mean}(\text{max}(|\bm{S}|))$.
    \item The identifiability score $\bm{c}=\text{mean}(\text{max}(|\bm{S_{adv}}|))$.
\end{itemize}
where the mean is over all data points and the max is taken over all output classes to capture the strongest feature contributions.

Once each feature is assigned a utility score ($v$) and an identifiability score ($c$), we can apply a greedy algorithm to solve the knapsack problem (as described in \S\ref{sec:game}).

\noindent \textbf{Greedy by utility.}
One approach to solving the problem is to prioritize utility by selecting features with the highest predictive power. We sort all features in descending order by their utility score $v_f$, and then iteratively select the top features until the utility constraint is met. A larger $V$ allows more features to be retained, maintaining higher accuracy at the cost of higher identifiability.

\noindent \textbf{Greedy by identifiability.} Another approach is to prioritize privacy by minimizing identifiability. We sort all features in ascending order by identifiability score $c_f$, and then iteratively choose the least identifiable features until the utility constraint is met. A larger $V$ allows more features to be selected, increasing both utility and identifiability.

\noindent \textbf{Greedy by cost-to-value ratio.} A more balanced approach is to consider the cost-to-value ratio, $c_f/v_f$, for each feature. Features are sorted in ascending order by this ratio, and the lowest-ratio features are selected iteratively until the utility constraint is met. This method strikes a balance between utility and privacy, prioritizing features that provide high predictive power with minimal identifiability risk. As before, $V$ remains a tunable parameter that controls the number of features selected.

\subsection{Hybrid Solution}\label{sec:hybrid}
Despite an efficient approach, the underlying assumption behind our greedy algorithm does not always hold.
There are cases where features interact with each other and can be correlated.
Selecting features independently based on their individual contributions can lead to suboptimal results.
To balance efficiency and accuracy, we propose a two-stage solution:
\begin{enumerate}
  \item Greedy Preselection (\S\ref{sec:greedy}): We first apply a greedy algorithm to reduce the feature set to a tractable size. \added[id=Design]{The user can specify a planned job duration for the exhaustive search, and the system automatically determines how many features to retain based on runtime estimates.}
  \item Exhaustive Search (\S\ref{sec:brute-force}): Within this reduced feature set, we then enumerate all possible combinations of the remaining features to identify the optimal subset.
\end{enumerate}
This hybrid approach leverages the speed of greedy selection to narrow down candidates while ensuring optimality within the reduced feature space. \S\ref{sec:exp-cost-benefit-exhaustive} evaluates how our hybrid solution is better than greedy itself.

%% file: evaluation.tex
\section{Evaluation}\label{sec:eval}

\begin{table}[]
\caption{\added[id=Organization]{Datasets used in experiments.}}\label{tab:dataset}
\small
\centering
\begin{tabular}{@{}lccc@{}}
\toprule
Dataset                        & \# classes & \# users & \# features \\ \midrule
Aposemat IoT-23~\cite{iot23-dataset}                & 2          & 31       & 9           \\
IoT Sentinel~\cite{miettinen2017iot}                   & 31         & 333      & 22          \\
Device Identification (NetML~\cite{netml})~\cite{shaowang2023amir} & 5         & 9      & 36         \\
CIC-IDS~\cite{cic-ids2017}                        & 2          & 271      & 74          \\
Opportunity~\cite{opportunity-dataset}                    & 5          & 4        & 135         \\ 
Device Identification (nPrint~\cite{nprint})~\cite{shaowang2023amir} & 5         & 9      & 2667         \\
Service Recognition (nPrint~\cite{nprint})~\cite{10.1145/3487552.3487842,10.1145/3366704,liu2024serveflow}  & 11         & 435      & 3285        \\
\bottomrule
\end{tabular}
\end{table}

\begin{table}[]
\caption{\added[id=Organization]{Comparisons between state-of-the-art and two-stage data minimization results.}}\label{tab:exp-teaser}
\small
\centering
\begin{tabular}{llllll}
\toprule
\multirow{2}{*}{Dataset}                & \multicolumn{2}{c}{SOTA} & \multicolumn{2}{c}{Feat. Min.} & \multirow{2}{*}{Rel. Eff.} \\
         & Acc.        & Ident.     & Acc.           & Ident.        \\ \midrule
Device Ident. (NetML)   & 95.62\%     & 69.52\%    & 94.78\% & 52.82\% & 2.990       \\
IoT Sentinel    & 70.60\%            & 62.03\%           & 69.96\%               & 50.86\% & 2.860             \\
CIC-IDS      & 97.98\%     & 57.68\%    & 97.02\%        & 45.85\%  & 2.511     \\
Aposemat IoT-23   & 99.91\%     & 77.98\%    & 99.05\%        & 69.81\% & 2.251      \\
Opportunity    & 87.77\%     & 99.87\%    & 81.68\%      & 66.46\%	 & 1.702 \\
Device Ident. (nPrint)   & 98.12\%     & 86.43\%    & 92.48\%        & 56.99\% & 1.652      \\
Service Recogn. (nPrint) & 95.52\%     & 91.63\%    & 84.95\% & 77.84\%    & 0.266   \\
\bottomrule
\end{tabular}
\end{table}

\begin{table*}[t]
\centering
\begin{minipage}{0.49\textwidth}
    \caption{Top 3 methods for each dataset.}\label{tab:top3-methods}
    \centering
    \footnotesize
    \begin{tabular}{lccc}
    \toprule
    Dataset & \emoji{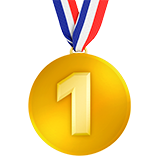} & \emoji{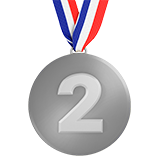} & \emoji{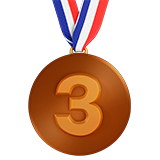}   \\ \midrule
    IoT Sentinel            & (8) SHAP by CTV & (3) Privacy & (4) Utility \\
    CIC-IDS 2017             & (1) Hashing & (4) Utility & (8) SHAP by CTV\\
    Opportunity              & (4) Utility & (5) Tradeoff & (8) SHAP by CTV\\
    Device Id. (NetML)    & (4) Utility & (6) SHAP by utility & (5) Tradeoff \\
    Device Id. (nPrint)   & (4) Utility & (3) Privacy & (5) Tradeoff\\
    Serv. Recog. (nPrint)       & (4) Utility & (5) Tradeoff & (3) Privacy\\
    \bottomrule
    \end{tabular}
\end{minipage}
\hfill
\begin{minipage}{0.24\textwidth}
    \centering
    \caption{IoT-23 (9 feats).}\label{tab:exp-exhaustive-iot23}
    \resizebox{\textwidth}{!}{
        \begin{tabular}{@{}$l^l^l^l^l@{}}
        \toprule
        Thr. & Acc. & Ident. & Rel. Eff. & \# Feat. \\ \midrule  
        0.0 & 99.91\% & 77.98\% & N/A & 5 \\
        \rowstyle{\bfseries}0.01 & 99.05\% & 69.81\% & 2.251 & 1 \\
        0.03 & 99.05\% & 69.81\% & 2.251 & 1 \\
        0.1 & 95.15\% & 60.60\% & 1.295 & 1 \\
        0.3 & 95.15\% & 60.60\% & 1.295 & 1 \\
        1.0 & 95.15\% & 60.60\% & 1.295 & 1 \\
        \bottomrule
        \end{tabular}
    }
\end{minipage}
\hfill
\begin{minipage}{0.24\textwidth}
    \centering
    \caption{IoT Sentinel (13 feats).}\label{tab:exp-exhaustive-iot-sentinel}
    \resizebox{\textwidth}{!}{
        \begin{tabular}{@{}$l^l^l^l^l@{}}
        \toprule
        Thr. & Acc. & Ident. & Rel. Eff. & \# Feat.  \\ \midrule  
        0.0 & 70.66\% & 62.02\% & N/A & 10 \\
        \rowstyle{\bfseries}0.01 & 69.96\% & 50.86\% & 2.769 & 8 \\
        0.03 & 69.03\% & 49.80\% & 2.014 & 4 \\
        0.1 & 63.61\% & 46.47\% & 0.791 & 4 \\
        0.3 & 49.93\% & 26.61\% & 0.535 & 2 \\
        1.0 & 21.34\% & 17.04\% & -0.092 & 1 \\
        \bottomrule
        \end{tabular}
    }
\end{minipage}

\hfill

\centering
\begin{minipage}{0.24\textwidth}
    \centering
    \caption{CIC-IDS (15 feats).}\label{tab:exp-exhaustive-cic-ids}
    \resizebox{\textwidth}{!}{
        \begin{tabular}{@{}$l^l^l^l^l@{}}
        \toprule
        Thr. & Acc. & Ident. & Rel. Eff. & \# Feat. \\ \midrule  
        0.0 & 97.85\% & 50.94\% & N/A & 5 \\
        \rowstyle{\bfseries} 0.01 & 97.02\% & 45.85\% & 1.814 & 2 \\
        0.03 & 95.14\% & 43.51\% & 1.009 & 2 \\
        0.1 & 91.10\% & 38.01\% & 0.650 & 1 \\
        0.3 & 70.70\% & 37.18\% & -0.680 & 1 \\
        1.0 & 70.70\% & 37.18\% & -0.680 & 1 \\
        \bottomrule
        \end{tabular}
    }
\end{minipage}
\hfill
\begin{minipage}{0.24\textwidth}
    \centering
    \caption{Opportunity (10 feats).}\label{tab:exp-exhaustive-opportunity}
    \resizebox{\textwidth}{!}{
        \begin{tabular}{@{}$l^l^l^l^l@{}}
        \toprule
        Thr. & Acc. & Ident. & Rel. Eff. & \# Feat. \\ \midrule  
        0.0 & 82.43\% & 69.21\% & N/A & 10 \\
        0.01 & 81.68\% & 66.46\% & 1.299 & 8 \\
        \rowstyle{\bfseries}0.03 & 80.10\% & 60.11\% & 1.362 & 7 \\
        0.1 & 75.40\% & 50.95\% & 0.955 & 5 \\
        0.3 & 58.39\% & 30.51\% & 0.476 & 3 \\
        1.0 & 42.76\% & 26.14\% & 0.082 & 2 \\
        \bottomrule
        \end{tabular}
    }
\end{minipage}
\begin{minipage}{0.24\textwidth}
    \centering
    \caption{Opportunity (7 feats).}\label{tab:exp-opportunity-7}
    \resizebox{\textwidth}{!}{
        \begin{tabular}{@{}$l^l^l^l^l@{}}
        \toprule
        Thr. & Acc. & Ident. & Rel. Eff. & \# Feat. \\ \midrule  
        0.0 & 78.91\% & 61.50\% & N/A & 7 \\
        0.01 & 78.31\% & 60.94\% & -0.069 & 6 \\
        0.03 & 76.71\% & 57.04\% & 0.707 & 6 \\
        0.1 & 74.16\% & 50.28\% & 0.860 & 4 \\
        0.3 & 58.53\% & 31.89\% & 0.374 & 3 \\
        1.0 & 42.74\% & 26.17\% & -0.023 & 1 \\
        \bottomrule
        \end{tabular}
    }
\end{minipage}
\hfill
\begin{minipage}{0.24\textwidth}
    \centering
    \caption{Opportunity (5 feats).}\label{tab:exp-opportunity-5}
    \resizebox{\textwidth}{!}{
        \begin{tabular}{@{}$l^l^l^l^l@{}}
        \toprule
        Thr. & Acc. & Ident. & Rel. Eff. & \# Feat. \\ \midrule 
        0.0 & 77.05\% & 57.39\% & N/A & 5 \\
        0.01 & 77.05\% & 57.39\% & N/A & 5 \\
        0.03 & 77.05\% & 57.39\% & N/A & 5 \\
        0.1 & 69.91\% & 49.40\% & 0.112 & 3 \\
        0.3 & 57.14\% & 33.51\% & 0.182 & 2 \\
        1.0 & 42.74\% & 26.17\% & -0.094 & 1 \\
        \bottomrule
        \end{tabular}
    }
\end{minipage}

\end{table*}

\subsection{Tasks and Datasets}
We selected 7 datasets to evaluate the approaches we introduced in earlier sections.
In all these tasks, there are a primary model and a user-classification (threat) model. The primary model makes predictions on the actual task the user cares about, whereas the threat model aims to infer the individual behind each request.
We use random forest models by default for both primary and threat models.
Table~\ref{tab:dataset} summarizes the datasets we used in experiments, \added[id=Organization]{and is sorted by the number of features.}
The number of classes refers to the possible outcomes of the primary model and the number of users means the possible outcomes of the threat model.
Out of the number of features shown in Table~\ref{tab:dataset}, our algorithm selects a subset of them to reach a trade-off between the primary model accuracy and the threat model identifiability.

\noindent \textbf{Network intrusion detection~\cite{cic-ids2017,iot23-dataset}.}
CIC-IDS 2017 and Aposemat IoT-23 are two network traffic datasets for intrusion detection analysis. The features are network traffic statistics such as duration, number of packets, number of bytes, length of packets, etc. The labels are binary (either benign or attack) but there are many possible IP addresses. We group them into IP subnets and use them as a proxy of users.
Our goal is to select features that are predictive for the intrusion detection task with the least IP leakage.

\noindent \textbf{Automated device-type identification for IoT~\cite{miettinen2017iot}.}
IoT Sentinel is a networking dataset for IoT device-type identification. \added[id=Design]{Its features are primarily derived from network headers and protocol metadata across all OSI layers, including protocol types, IP options, port usage, and packet sizes. Each sample is labeled with a specific device type, such as Hue Switch or TP-Link Plug.} Compared to CIC-IDS 2017, IoT Sentinel includes fewer features but encompasses a broader range of device classes. As a result, the primary model achieves lower accuracy than models trained on CIC-IDS 2017.

\noindent \textbf{Sensor-based human activity recognition~\cite{opportunity-dataset}.}
Opportunity is a sensor-based dataset for human activity recognition.
We use locomotion prediction (sit, walk, stand, lie) as the primary model and user prediction as the threat model.

\noindent \textbf{Device identification~\cite{shaowang2023amir}.}
We leverage the open-source dataset from AMIR~\cite{shaowang2023amir} covering 5 in-home IoT devices.
\added[id=Design]{The label schema is similar to that of IoT Sentinel, but the dataset provides two distinct feature representations: a dense format from NetML~\cite{netml} and a sparse format from nPrint~\cite{nprint}.} A detailed ablation study comparing these two representations is presented in \S\ref{sec:choice-featurization}.

\noindent \textbf{Service recognition~\cite{10.1145/3487552.3487842,10.1145/3366704,liu2024serveflow}.}
We aggregate three datasets of 23,487 flows in total as our service recognition task. \added[id=Design]{The labels correspond to network application types, including video conferencing (e.g., Zoom, Google Meet), video streaming (e.g., YouTube, Twitch), and social media (e.g., Instagram, Facebook).}
In our primary model, those combined flows are categorized into 11 classes of applications. This dataset only comes in nPrint~\cite{nprint} sparse representations.




\subsection{Ranking Feature Selection Methods}
\added[id=Clarification]{For greedy selection,} we employ the following feature selection methods on aforementioned datasets: (1) feature hashing, (2) PCA, (3) entropy-based privacy score, (4) mutual information-based utility score, (5) privacy-utility tradeoff score, (6) SHAP-based greedy by utility, (7) SHAP-based greedy by identifiability, (8) SHAP-based greedy by cost-to-value ratio, (9) Gini impurity-based greedy by utility, (10) Gini impurity-based greedy by identifiability, (11) Gini impurity-based greedy by cost-to-value ratio, and (12) differential privacy. Detailed description of each method can be found in \S\ref{sec:baselines}.

We rank these methods based on the highest relative effectiveness, and Table~\ref{tab:top3-methods} shows the top 3 methods for each dataset. Although there is no clear winner, the best performing methods are (4) utility score and (8) SHAP-based greedy by CTV across different datasets.
Traditional feature selection methods, such as PCA, do not perform well.
For more detailed comparisons between feature selection methods, a case study can be found in \S\ref{sec:exp-case-study}.

\added[id=Clarification]{Taking exhaustive search into account as well,} Table~\ref{tab:exp-teaser} shows a summary of our \added[id=Clarification]{two-stage} data minimization results, \added[id=Organization]{and is sorted by relative effectiveness.} The base accuracy and identifiability for relative effectiveness come from the full dataset without data minimization.
10-15 features are selected from the greedy stage by SHAP or utility score, and 0.01 is used as the threshold in the exhaustive search stage.

\subsection{Effectiveness of Exhaustive Search}
\subsubsection{Exhaustive search works well for a dataset with a small number of features.}
Given a dataset with tractable dimensionality, our exhaustive search algorithm described in \S\ref{sec:brute-force} provides the optimal accuracy-identifiability tradeoff by enumerating all possible feature subsets. On the Aposemat IoT-23 dataset (Table~\ref{tab:exp-exhaustive-iot23}), we find that retaining only the ``orig\_pkts'' feature on cleaned data (without NaNs) achieves an 8\% reduction in identifiability with less than 1\% accuracy loss.
If we relax the threshold, another feature is chosen and the identifiability can further be reduced to 60.60\% while maintaining 95\% primary model accuracy.

\subsubsection{For larger datasets with dense features, our two-stage hybrid solution achieves good tradeoffs.}
When the dataset contains a larger number of features, exhaustive search becomes computationally infeasible due to its exponential complexity.
In these cases, we resort to the greedy algorithm (\S\ref{sec:greedy}) to reduce the feature space, followed by exhaustive search on the remaining features.
Table~\ref{tab:exp-exhaustive-iot-sentinel},~\ref{tab:exp-exhaustive-cic-ids},~\ref{tab:exp-exhaustive-opportunity} illustrate how our \added[id=Clarification]{two-stage} approach achieves a favorable tradeoff between model accuracy and identifiability across various telemetry-based datasets. Here, the base accuracy and identifiability for relative effectiveness come from a 0-threshold exhaustive search after the greedy stage. Empirically, setting the threshold to 0.01 appears to be a reasonable choice across multiple datasets.


\begin{figure*}[htbp]
    \centering
    \begin{subfigure}{0.24\linewidth}
        \centering
        \includegraphics[width=\linewidth]{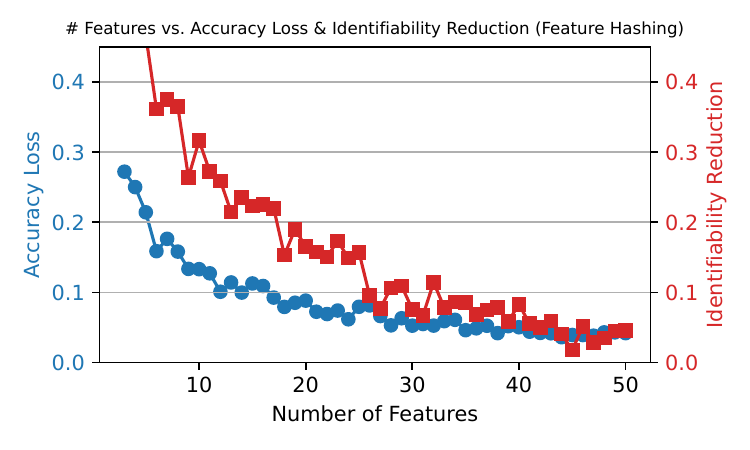}
        \caption{Feature hashing.}
        \label{fig:opportunity-acc-ident-a}
    \end{subfigure}
    \hfill
    \begin{subfigure}{0.24\linewidth}
        \centering
        \includegraphics[width=\linewidth]{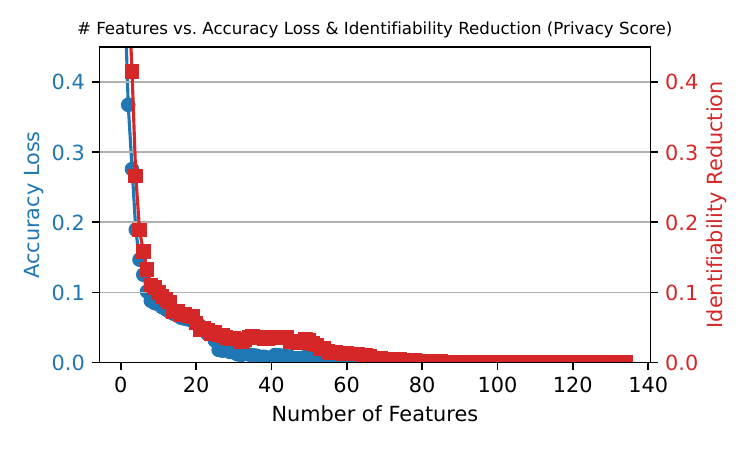}
        \caption{Privacy score.}
        \label{fig:opportunity-acc-ident-c}
    \end{subfigure}
    \hfill
    \begin{subfigure}{0.24\linewidth}
        \centering
        \includegraphics[width=\linewidth]{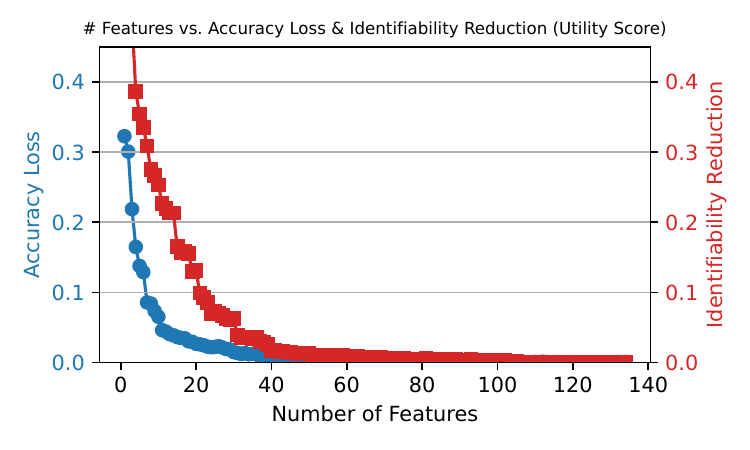}
        \caption{Utility score.}
        \label{fig:opportunity-acc-ident-d}
    \end{subfigure}
    \hfill
    \begin{subfigure}{0.24\linewidth}
        \centering
        \includegraphics[width=\linewidth]{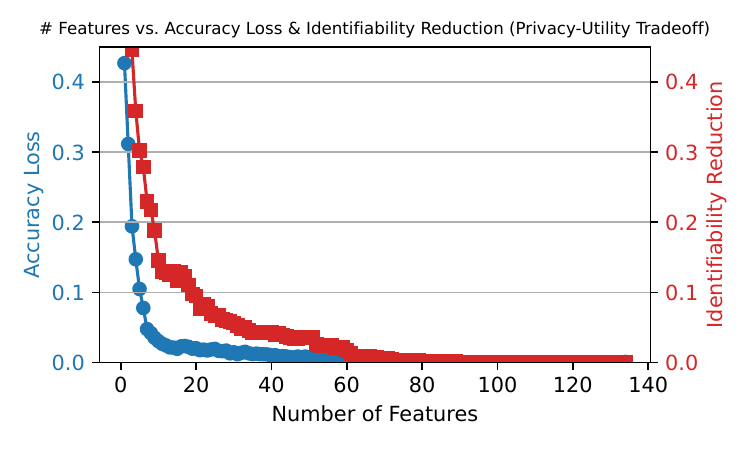}
        \caption{Privacy-utility tradeoff.}
        \label{fig:opportunity-acc-ident-e}
    \end{subfigure}
    \begin{subfigure}{0.24\linewidth}
        \centering
        \includegraphics[width=\linewidth]{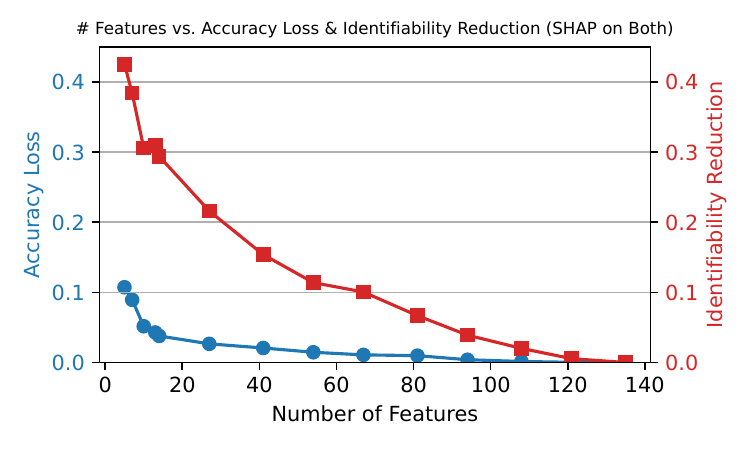}
        \caption{SHAP greedy by CTV.}
        \label{fig:opportunity-acc-ident-f}
    \end{subfigure}
    \hfill
    \begin{subfigure}{0.24\linewidth}
        \centering
        \includegraphics[width=\linewidth]{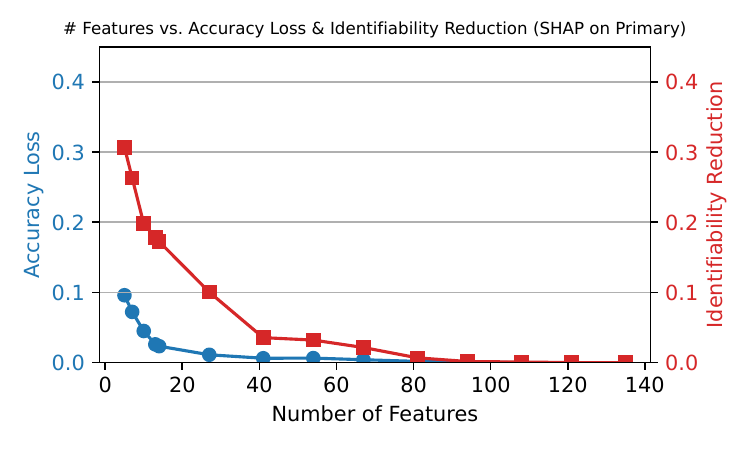}
        \caption{SHAP greedy by utility.}
        \label{fig:opportunity-acc-ident-g}
    \end{subfigure}
    \hfill
    \begin{subfigure}{0.24\linewidth}
        \centering
        \includegraphics[width=\linewidth]{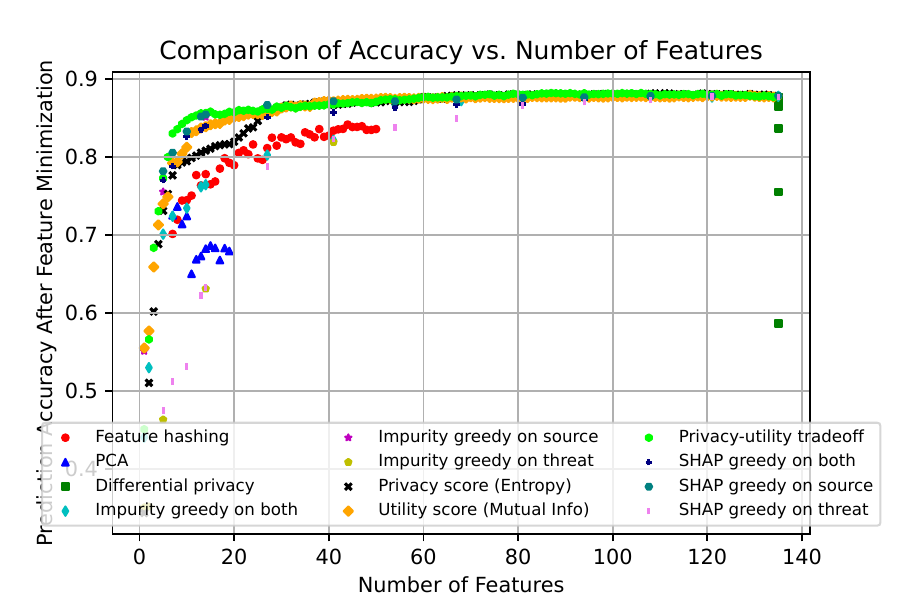}
        \caption{Accuracy vs. \# features.}
        \label{fig:opportunity-accuracy-features}
    \end{subfigure}
    \hfill
    \begin{subfigure}{0.24\linewidth}
        \centering
        \includegraphics[width=\linewidth]{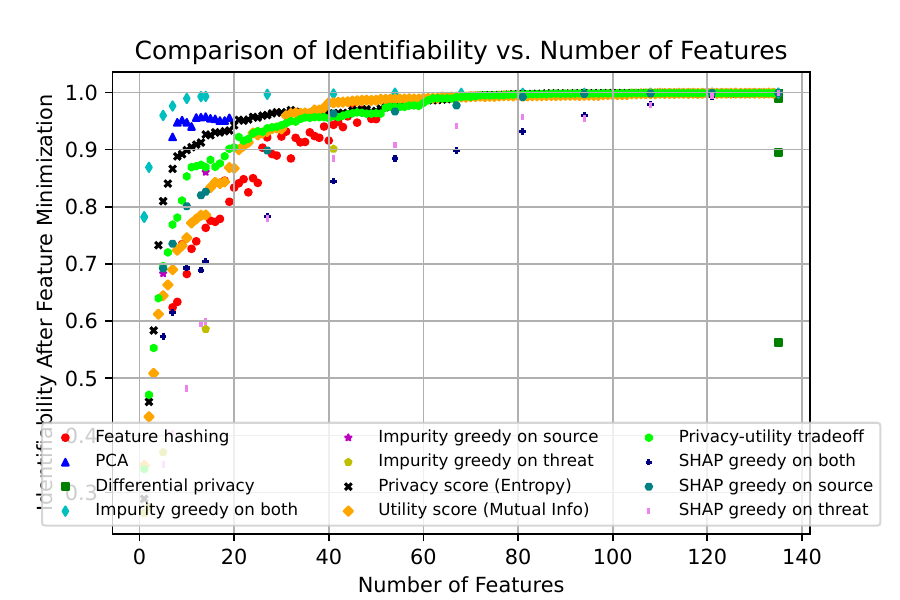}
        \caption{Identifiability vs. \# features.}
        \label{fig:opportunity-ident-features}
    \end{subfigure}
    \caption{Case study: Opportunity dataset (greedy selection).}
    \label{fig:opportunity}
\end{figure*}

\begin{table}[tbp]
\centering
\begin{minipage}{0.49\columnwidth}
    \centering
    \caption{Dev. Ident. (NetML)}\label{tab:exp-exhaustive-device-ident-netml}
    \resizebox{\textwidth}{!}{
        \begin{tabular}{@{}$l^l^l^l^l@{}}
        \toprule
        Thr. & Acc. & Ident. & Rel. Eff. & \# Feat. \\ \midrule  
        0.0 & 95.62\% & 56.37\% & N/A & 4 \\
        \rowstyle{\bfseries}0.01 & 94.78\% & 52.82\% & 1.441 & 3 \\
        0.03 & 93.53\% & 49.27\% & 1.223 & 4 \\
        0.1 & 86.22\% & 42.17\% & 0.413 & 3 \\
        0.3 & 73.90\% & 34.03\% & 0.028 & 1 \\
        1.0 & 73.90\% & 34.03\% & 0.028 & 1 \\
        \bottomrule
        \end{tabular}
    }
\end{minipage}
\hfill
\begin{minipage}{0.49\columnwidth}
    \centering
    \caption{Dev. Ident. (nPrint)}\label{tab:exp-exhaustive-device-ident-nprint}
    \resizebox{\textwidth}{!}{
        \begin{tabular}{@{}$l^l^l^l^l@{}}
        \toprule
        Thr. & Acc. & Ident. & Rel. Eff. & \# Feat. \\ \midrule  
        0.0 & 93.32\% & 59.71\% & N/A & 10 \\
        \rowstyle{\bfseries}0.01 & 92.48\% & 56.99\% & 1.175 & 7 \\
        0.03 & 90.61\% & 54.49\% & 0.656 & 8 \\
        0.1 & 84.34\% & 47.60\% & 0.299 & 7 \\
        0.3 & 66.81\% & 36.53\% & -0.134 & 2 \\
        1.0 & 46.76\% & 31.52\% & -0.502 & 1 \\
        \bottomrule
        \end{tabular}
    }
\end{minipage}
\hfill
\begin{minipage}{0.49\columnwidth}
    \centering
    \caption{Serv. Recog. (nPrint)}\label{tab:exp-exhaustive-service-recognition}  
    \resizebox{\textwidth}{!}{
        \begin{tabular}{@{}$l^l^l^l^l@{}}
        \toprule
        Thr. & Acc. & Ident. & Rel. Eff. & \# Feat. \\ \midrule  
        0.0 & 85.65\% & 78.83\% & N/A & 10 \\
        \rowstyle{\bfseries}0.01 & 84.95\% & 77.84\% & 0.347 & 8 \\
        0.03 & 83.27\% & 75.94\% & 0.194 & 8 \\
        0.1 & 77.16\% & 71.48\% & -0.144 & 5 \\
        0.3 & 60.33\% & 57.39\% & -0.166 & 3 \\
        1.0 & 37.60\% & 40.65\% & -0.230 & 1 \\
        \bottomrule
        \end{tabular}
    }
\end{minipage}
\hfill
\begin{minipage}{0.49\columnwidth}
    \centering
    \caption{Opp. (10 feats, 50\%)}\label{tab:exp-exhaustive-opportunity-0.5}
    \added[id=Design]{
    \resizebox{\textwidth}{!}{
        \begin{tabular}{@{}$l^l^l^l^l@{}}
        \toprule
        Thr. & Acc. & Ident. & Rel. Eff. & \# Feat. \\ \midrule  
        0.0 & 81.24\% & 65.31\% & N/A & 10 \\
        0.01 & 80.56\% & 62.54\% & 1.405 & 9 \\
        \rowstyle{\bfseries}0.03 & 79.79\% & 59.13\% & 1.450 & 8 \\
        0.1 & 73.15\% & 48.39\% & 0.738 & 4 \\
        0.3 & 58.10\% & 30.03\% & 0.422 & 3 \\
        1.0 & 42.23\% & 25.77\% & 0.013 & 1 \\
        \bottomrule
        \end{tabular}
    }
    }
\end{minipage}
\hfill
\begin{minipage}{0.49\columnwidth}
    \centering
    \caption{Opp. (10 feats, 10\%)}\label{tab:exp-exhaustive-opportunity-0.1}
    \added[id=Design]{
    \resizebox{\textwidth}{!}{
        \begin{tabular}{@{}$l^l^l^l^l@{}}
        \toprule
        Thr. & Acc. & Ident. & Rel. Eff. & \# Feat. \\ \midrule  
        0.0 & 76.74\% & 50.95\% & N/A & 8 \\
        \rowstyle{\bfseries}0.01 & 76.09\% & 48.73\% & 1.228 & 8 \\
        0.03 & 75.10\% & 45.53\% & 1.195 & 7 \\
        0.1 & 71.32\% & 40.34\% & 0.672 & 6 \\
        0.3 & 56.13\% & 28.61\% & 0.081 & 3 \\
        1.0 & 37.39\% & 25.60\% & -0.440 & 1 \\
        \bottomrule
        \end{tabular}
    }
    }
\end{minipage}
\hfill
\begin{minipage}{0.49\columnwidth}
    \centering
    \caption{Opp. (MLP, 10 feats)}\label{tab:exp-exhaustive-opportunity-mlp}
    \added[id=Design]{
    \resizebox{\textwidth}{!}{
        \begin{tabular}{@{}$l^l^l^l^l@{}}
        \toprule
        Thr. & Acc. & Ident. & Rel. Eff. & \# Feat. \\ \midrule  
        0.0 & 77.71\% & 53.34\% & N/A & 7 \\
        \rowstyle{\bfseries}0.01 & 77.08\% & 49.75\% & 1.740 & 8 \\
        0.03 & 75.44\% & 45.49\% & 1.241 & 6 \\
        0.1 & 70.40\% & 36.27\% & 0.848 & 4 \\
        0.3 & 60.50\% & 28.28\% & 0.376 & 3 \\
        1.0 & 46.90\% & 26.49\% & -0.138 & 1 \\
        \bottomrule
        \end{tabular}
    }
    }
\end{minipage}
\hfill
\begin{minipage}{0.49\columnwidth}
    \centering
    \caption{Opp. (KNN, 10 feats)}\label{tab:exp-exhaustive-opportunity-knn}
    \added[id=Design]{
    \resizebox{\textwidth}{!}{
        \begin{tabular}{@{}$l^l^l^l^l@{}}
        \toprule
        Thr. & Acc. & Ident. & Rel. Eff. & \# Feat. \\ \midrule  
        0.0 & 78.32\% & 63.62\% & N/A & 7 \\
        \rowstyle{\bfseries}0.01 & 77.57\% & 59.21\% & 1.772 & 9 \\
        0.03 & 76.43\% & 54.66\% & 1.556 & 7 \\
        0.1 & 72.51\% & 47.23\% & 1.037 & 5 \\
        0.3 & 56.19\% & 27.21\% & 0.498 & 2 \\
        1.0 & 40.45\% & 25.48\% & 0.007 & 1 \\
        \bottomrule
        \end{tabular}
    }
    }
\end{minipage}
\hfill
\begin{minipage}{0.49\columnwidth}
    \centering
    \includegraphics[width=\linewidth]{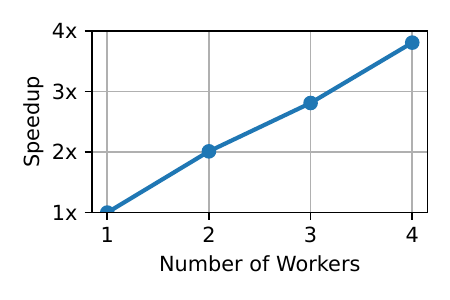}
    \captionof{figure}{Scalability.}
    \label{fig:opportunity-scalability}
\end{minipage}
\end{table}

\subsection{Ablation Study}\label{sec:ablation}
The purpose of our greedy algorithm (\S\ref{sec:greedy}) is to reduce the number of features to at most 10--15 while retaining as much information as possible.
Ideally, the reduced feature subset should be informative enough to allow a subsequent exhaustive search to find the optimal tradeoff between accuracy and identifiability.



\subsubsection{Reducing the number of features in exhaustive search will lead to worse tradeoffs}\label{sec:exp-cost-benefit-exhaustive}
Since exhaustive search is so computationally expensive, why not rely solely on greedy selection?
As shown in Fig.~\ref{fig:opportunity-acc-ident-d}, a substantial accuracy loss is often observed when the number of features retained by greedy preselection is small. This sudden drop is why we adopt exhaustive search as a second stage, allowing fine-grained optimization of the selected feature subset.

To assess the impact of greedy preselection on the exhaustive search stage, we vary the number of features retained by the greedy algorithm and compare the corresponding performance in \added[id=Clarification]{our two-stage solution}.
By examining Table~\ref{tab:exp-exhaustive-opportunity}, Table~\ref{tab:exp-opportunity-7}, and Table~\ref{tab:exp-opportunity-5}, we observe that the best tradeoff between accuracy and identifiability is achieved when 10 features are retained by the greedy preselection stage, compared to 5 or 7 features. This is because retaining more features allows for greater flexibility in the subsequent exhaustive search, enabling it to find an optimal subset that balances accuracy and privacy more effectively.
However, retaining more features means that more computation is required for the exhaustive search.


\subsubsection{Dense feature representations lead to better tradeoffs than sparse representations.}\label{sec:choice-featurization}
For the device identification dataset, we evaluated two different featurization of the same data: NetML~\cite{netml} and nPrint~\cite{nprint}.
NetML is a featurized representation with 36 dense features, where each feature captures higher-level statistical properties of network traffic.
nPrint, in contrast, represents the raw bit-level packet data with 2667 sparse features, where each feature can take values 1, 0, or -1 (with -1 indicating non-existent headers in a given packet).
Unlike dense feature representations, sparse feature sets distribute information across a large number of low-information features, meaning that each individual feature carries limited predictive value.
\added[id=Evaluation]{We find that our two-stage method performs well on the NetML representation but is less effective on nPrint's sparse representation (Table~\ref{tab:exp-exhaustive-device-ident-netml}, Table~\ref{tab:exp-exhaustive-device-ident-nprint}).}

Since the complexity of exhaustive search grows exponentially with the number of features retained by greedy selection, sparse feature sets present a challenge: less information is contained in the same number of features compared to dense representations. As a result, even though exhaustive search attempts to find the optimal subset, it is less effective in sparse settings because the optimization space is inherently limited.
This limitation is more evident in the service recognition dataset (Table~\ref{tab:exp-exhaustive-service-recognition}), where relative effectiveness remains low despite reductions in both accuracy and identifiability. This is expected, as \textit{sparse features tend to contribute to both accuracy and identifiability simultaneously. They are both predictive and revealing, making the tradeoff less favorable.}
Thus, our \added[id=Clarification]{two-stage method} performs better on datasets with dense features, where more meaningful tradeoffs between identifiability and accuracy can be achieved.

While in theory, a better utility-identifiability tradeoff could be found using sparse feature representations like nPrint given unlimited computational resources, the sheer number of possible feature subsets makes exhaustive search impractical. The potential improvement is unlikely to justify the additional computational cost, making dense feature representations a more practical choice for data minimization in real-world applications.

\subsubsection{Our method scales with both the number of rows and workers.}
\added[id=Design]{
Table~\ref{tab:exp-exhaustive-opportunity-0.5} and Table~\ref{tab:exp-exhaustive-opportunity-0.1} follow the same two-stage experimental setup as Table~\ref{tab:exp-exhaustive-opportunity}, with the only difference being that 50\% and 10\% of the data, respectively, are used to train both the primary and threat models. While both accuracy and identifiability decrease slightly, the drop is not substantial, and our method still achieves a reasonable level of relative effectiveness.}

\added[id=Design]{
As described in \S\ref{sec:practical-solutions}, the exhaustive search is the primary computational bottleneck, while the greedy solution is relatively fast to run. Fig.~\ref{fig:opportunity-scalability} illustrates the scalability of exhaustive search on the Opportunity dataset.
Since the search space can be easily partitioned by distributing feature subsets evenly across workers, our approach achieves near-linear speedup through parallelization.
}

\subsubsection{Our method is transferrable across different models.}
\added[id=Design]{
In all aforementioned experiments, we have used random forest as both the primary and threat models for consistency.
To assess the robustness of our findings across different model architectures, Table~\ref{tab:exp-exhaustive-opportunity-mlp} and Table~\ref{tab:exp-exhaustive-opportunity-knn} present results from applying our two-stage data minimization approach using multi-layer perceptron (MLP) and k-nearest neighbors (KNN), respectively, on the same Opportunity dataset.
The results are largely consistent with those obtained using random forest, indicating that our conclusions generalize well across different model choices.
}

\subsection{Case Study: Opportunity Dataset}\label{sec:exp-case-study}
Fig.~\ref{fig:opportunity} presents a comprehensive analysis of \added[id=Clarification]{various greedy selection methods} applied to the Opportunity human activity recognition dataset.
The key observations are summarized below.

\subsubsection{Identifiability reduction exceeds accuracy loss.}
In Fig.~\ref{fig:opportunity-acc-ident-a}--\ref{fig:opportunity-acc-ident-g}, we compare identifiability reduction (right y-axis) against accuracy loss (left y-axis) across various feature selection methods. The accuracy loss (blue line with rounded dots) is generally lower than the identifiability reduction (red line with squared dots) for a given number of selected features.
This indicates that our feature selection methods effectively reduce identifiability while keeping accuracy degradation minimal, validating our approach for privacy-aware data minimization.

\subsubsection{Spike in accuracy loss when too few features are retained.}
When the number of features becomes too small, both accuracy loss and identifiability reduction increase sharply, as shown in Fig.~\ref{fig:opportunity-acc-ident-a}--\ref{fig:opportunity-acc-ident-g},~\ref{fig:opportunity-accuracy-features},~\ref{fig:opportunity-ident-features}.
This is problematic because accuracy loss at this level is typically unacceptable, reinforcing the need to perform exhaustive search for finer-grained optimization with a reasonable number of features.
Across all feature selection methods, performance becomes less stable when the number of features is reduced too aggressively, highlighting the importance of our two-stage hybrid solution.

\subsubsection{Limitations of relative effectiveness.}\label{sec:rel-eff-limitations}
While relative effectiveness is a useful metric for evaluating identifiability reduction per unit of accuracy loss, it has two key limitations:
    (1) The metric can be amplified when accuracy loss is negligible, making the tradeoff appear more favorable than it actually is.
    (2) Even when relative effectiveness appears high, both accuracy and identifiability may have declined significantly, potentially leading to unacceptable accuracy for practical use.

\added[id=Clarification]{In Fig.~\ref{fig:opportunity-acc-ident-d}, relative effectiveness is the highest when 103 features are selected. However, both accuracy drop and identifiability reduction at that point are minimal, which inflates the metric. Conversely, although the left tail (i.e., when few features are retained) may still yield seemingly reasonable relative effectiveness, the accompanying accuracy degradation is often too severe to be acceptable.}

\subsection{Recommendations for Best Practices}
Based on our experiments, we recommend a structured approach for effective data minimization:

\noindent \textbf{Choose the right feature representations: prioritize dense representations over sparse alternatives.}
When initiating data minimization, dense representations (e.g., NetML) are preferred over sparse representations (e.g., nPrint). Dense features inherently encode more information per feature, enabling computationally efficient optimization while effectively reducing identifiability. This contrasts with sparse representations, which often require retaining larger feature sets for utility, conflicting with minimization goals.

\noindent \textbf{Pre-select features: use mutual information and SHAP-based methods.}
Mutual information-based feature utility scores and SHAP-based greedy selection using cost-to-value ratio (CTV) are two effective methods for preliminary data minimization. Our findings also indicate that traditional feature selection methods (e.g., PCA, feature hashing) \textit{should be avoided for privacy-aware feature selection.}
Depending on the dataset size and available computational resources, retaining \textit{10–15 features} is recommended to achieve better overall relative effectiveness in the next stage.

\noindent \textbf{Minimize identifiability by exhaustive search: a 0.01 threshold offers a practical tradeoff.}
\added[id=Evaluation]{In our experiments, setting the accuracy loss tolerance parameter $l=0.01$ appears to achieve a favorable balance between predictive performance and user identifiability in most cases. This threshold allows for meaningful reductions in identifiability while keeping the impact on predictive accuracy minimal. While other applications may tolerate larger accuracy losses, we find that a small accuracy drop (e.g., 1\%) is generally acceptable and aligns with the overarching goal of maintaining functional model performance.}



%% file: discussion.tex
\section{Discussion and Future Work}\label{sec:disc}

Table~\ref{tab:top3-methods} confirms that neither feature hashing nor PCA improves performance when applied to sparse features (e.g., nPrint~\cite{nprint}).
This is because each coordinate in the nPrint vector merely records the presence of a single packet bit; projecting such sparse binary vectors does not introduce new, semantically rich signals. In this context, ``densification'' functions more as a compression technique than as a method of feature selection.
A straightforward remedy is to re-encode the same traffic with a denser feature set, such as the flow-level statistical vectors provided by NetML~\cite{netml}, where each attribute aggregates many bytes and thus carries richer variation.

Beyond manual feature engineering, foundation models for packet data now offer a direct path to densification: netFound~\cite{guthula2023netfound} learns multimodal embeddings from large unlabeled traces, and NetLLM~\cite{wu2024netllm} shows that such embeddings can be fine-tuned quickly for downstream intrusion-detection and classification tasks. Mapping nPrint outputs into these learned embedding spaces would supply the dense features that our results suggest are necessary.